\documentclass[journal]{IEEEtran}
\usepackage{cite}
\usepackage{amsmath,amssymb,amsfonts}
\usepackage{algorithmic}
\usepackage{graphicx}
\usepackage{textcomp}
\usepackage{url}
\usepackage{tabularx}
\usepackage{multirow}
\usepackage{CJKutf8} 
\usepackage{array}
\usepackage{booktabs}
\usepackage{hhline}
\usepackage{diagbox}
\usepackage{caption}
\usepackage[utf8]{inputenc}
\usepackage[T1]{fontenc}
\usepackage{tabularray}
\usepackage[colorlinks=true, allcolors=blue]{hyperref}
\usepackage{xcolor}
\usepackage{geometry}

\begin{document}
\title{Fine-grained Speech Sentiment Analysis in Chinese Psychological Support Hotlines Based on Large-scale Pre-trained Model}
\author{Zhonglong Chen, Changwei Song, Yining Chen, Jianqiang Li, Guanghui Fu, Yongsheng Tong\textsuperscript{*} and Qing Zhao\textsuperscript{*}
\thanks{Zhonglong Chen, Changwei Song, Yining Chen, Jianqiang Li, Qing Zhao are with School of Software Engineering, Beijing University of Technology, Beijing, China.}
\thanks{Yongsheng Tong is with Peking University Huilongguan Clinical Medical School, Beijing, China; WHO Collaborating Center for Research and Training in Suicide Prevention, Beijing, China.}
\thanks{Guanghui Fu is with Sorbonne Universit\'{e}, Institut du Cerveau – Paris Brain Institute - ICM, CNRS, Inria, Inserm, AP-HP, H\^{o}pital de la Piti\'{e}-Salp\^{e}tri\`{e}re, Paris, France (e-mail: guanghui.fu@inria.fr).}
\thanks{Corresponding authors: Qing Zhao (\url{zhaoqing@bjut.edu.cn}) and Yongshen Tong (\url{timystong@pku.org.cn}).}
\thanks{This work was supported by The National Natural Science Foundation of China under Grant No.82071546.}
\thanks{This work has been submitted to the IEEE for possible publication. Copyright may be transferred without notice, after which this version may no longer be accessible.}
}
\maketitle
\begin{abstract}
Suicide and suicidal behaviors remain significant challenges for public policy and healthcare. In response, psychological support hotlines have been established worldwide to provide immediate help to individuals in mental crises. The effectiveness of these hotlines largely depends on accurately identifying callers' emotional states, particularly underlying negative emotions indicative of increased suicide risk. However, the high demand for psychological interventions often results in a shortage of professional operators, highlighting the need for an effective speech emotion recognition model. This model would automatically detect and analyze callers' emotions, facilitating integration into hotline services. Additionally, it would enable large-scale data analysis of psychological support hotline interactions to explore psychological phenomena and behaviors across populations.
Our study utilizes data from the Beijing psychological support hotline, the largest suicide hotline in China. We analyzed speech data from 105 callers containing 20,630 segments and categorized them into 11 types of negative emotions. We developed a negative emotion recognition model and a fine-grained multi-label classification model using a large-scale pre-trained model. Our experiments indicate that the negative emotion recognition model achieves a maximum F1-score of 76.96\%. However, it shows limited efficacy in the fine-grained multi-label classification task, with the best model achieving only a 41.74\% weighted F1-score. We conducted an error analysis for this task, discussed potential future improvements, and considered the clinical application possibilities of our study. All the codes are public available at: \url{https://github.com/czl0914/psy_hotline_analysis}.
\end{abstract}
\IEEEpeerreviewmaketitle

\section{Introduction}
\IEEEPARstart{T}{he} growing global burden and social impact of suicide highlight the critical need for effective public health interventions. Annually, approximately 817,000 people die by suicide worldwide, making it the second leading cause of premature death among young people~\cite{naghavi2019global}. Despite a decline in the overall suicide rate in China, the incidence of attempted suicide remains alarmingly high at about 0.8\%~\cite{cao2015prevalence}. This trend is particularly concerning among adolescents, with suicide rates continuing to rise each year~\cite{chen2018suicidal}, emphasizing the urgent need for effective preventive measures. In response, many countries, ranging from developed to developing nations, have established nationwide networks of psychological support hotlines. These hotlines have proven to be an important and highly effective tool for suicide prevention~\cite{gould2018follow, gould2016helping}, connecting individuals in crisis with trained professionals who provide immediate telephone counseling, comprehensive risk assessment, and necessary interventions~\cite{witte2010assessing, zhao2021comparisons}. Callers to psychological support hotlines typically express a range of negative emotions in their speech~\cite{shaw2019taiwan, ramchand2016characteristics}. Accurately recognizing the emotional state of callers can enable operators to quickly analyze and continuously monitor changes in emotional states during calls~\cite{gould2018follow}, which is highly beneficial for psychological counseling. 
Tavi et al.~\cite{tavi2020phonetic} studied the prosodic changes in emergency suicide calls. Wang et al.~\cite{wang2023self} based on actual phone recordings from Taiwan's Lifeline, used NLP techniques to select statements that clearly show suicide risk signals, attempting to enhance the detection capabilities of suicidal ideation within suicide prevention hotlines. Iyer and Ravi~\cite{iyer2023unified} explored methods to detect suicide risk through text and voice signal analysis based on recordings from Australia's national suicide prevention hotline service. Salmi et al.~\cite{salmi2022detecting} used BERTopic~\cite{grootendorst2022bertopic} to analyze the changes in conversation content on suicide prevention hotlines during the COVID-19 pandemic. Alabdulla et al.~\cite{alabdulla2023management} studied how the Qatar National Mental Health Helpline managed suicide and self-harm risks during the COVID-19 pandemic. These studies vary in method and focus but underscore the importance of data-driven analysis in enhancing the efficiency of suicide prevention.
However, research that focuses on identifying the emotional states of suicide hotline callers, especially fine-grained negative emotional states, remains scarce. In particular, there are few studies of fine-grained emotion recognition using Chinese-language suicide hotline speech data. 

In recent years, the field of artificial intelligence has witnessed significant advancements through the development of large-scale pre-trained models (PTMs) containing hundreds of millions, or even billions, of parameters, which have excelled in a variety of tasks~\cite{liu2022audio,du2022survey}. PTMs, capable of handling multiple tasks simultaneously, are pre-trained on extensive amounts of unlabeled data and then fine-tuned for specific tasks, often achieving human-like performance~\cite{brown2020language}. Prominent models such as BERT~\cite{devlin2018bert}, GPT-3~\cite{brown2020language}, and CLIP~\cite{radford2021learning} have demonstrated exceptional results in natural language processing and computer vision tasks, particularly gaining significant attention in the field of psychology~\cite{he2023towards}.

In the realm of speech signal processing, models like Wav2Vec~\cite{schneider2019wav2vec}, Wav2Vec 2.0~\cite{baevski2020wav2vec}, HuBERT~\cite{hsu2021hubert}, and Whisper~\cite{whisper_radford2023robust} have been based on large-scale, unsupervised speech training and have shown impressive capabilities in speech-related tasks, achieving satisfactory performance. However, Speech Emotion Recognition (SER) encounters unique challenges in the healthcare domain, particularly in psychological support hotlines. Callers with suicidal ideation often exhibit speech characteristics such as slow speech rates and low volume~\cite{tavi2020phonetic}, necessitating models that can recognize these subtle differences. Despite the potential, there is a noticeable gap in clinical studies, and the emotion recognition performance of these pre-trained models in psychological support hotlines remains unexplored.

In this study, we developed our models using data from China's largest psychological support hotline, based at Beijing Huilongguan Hospital. We selected voice data from 105 callers, comprising 20,630 segments, which were annotated with positive and negative emotions by three professionals. Additionally, we defined eleven detailed fine-grained emotion categories. We constructed both a binary emotion classification model (positive vs. negative) and a fine-grained multi-label emotion classification model using large-scale pre-trained models. 
The experimental results revealed that the performance of several pre-trained models in the binary classification task was comparable, with the Wav2Vec 2.0 model \cite{baevski2020wav2vec} achieving the highest F1-score of 76.96\%. However, these models performed less effectively in the fine-grained emotion recognition task, with the best-performing model, Whisper \cite{whisper_radford2023robust}, achieving a weighted F1-score of only 41.74\%. 

To our knowledge, this study is the first to conduct fine-grained emotion recognition in a Chinese-language psychiatric helpline setting. Our study parallels that of Nfissi et al. \cite{nfissi2024unlocking}, with a more detailed task definition (11 categories compared to their 4) and a larger sample size (20,630 segments compared to their 999). Our research is pioneering and has significant clinical implications. It can serve as a valuable tool for integration into psychological support hotlines systems and for large-scale population psychometric analysis.


\section{Datasets} \label{sec:datasets} 
Our experimental data were sourced from the Beijing Psychological Intervention Hotline at Huilongguan Hospital in Beijing, which is the first and largest suicide hotline in China. From this resource, we randomly selected 105 sets of recorded conversations between operators and callers, resulting in a total of 20,630 segmented caller voice sentences. Three experts labeled these segments for negative emotions across 11 fine-grained categories, and the final labels were chosen as a concatenation of the labeling results of the three experts. The dataset includes 9,774 segments identified with negative emotions and 10,856 segments with non-negative emotions. Of the segments with negative emotions, ``Sadness'' was the most prevalent, accounting for 5,662 entries, making it the dominant emotion category. This data distribution for fine-grained sentiment analysis is detailed in Table~\ref{tab:data_fine_grained}. 

\begin{table}[!ht]
\centering
\caption{Data distribution for fine-grained sentiment multi-label classification tasks.}
\label{tab:data_fine_grained}
\begin{tabular}{ll} 
\hline
Categories   & Segments  \\ 
\hline
1. Sadness      & 5662      \\
2. Pain         & 2102      \\
3. Grievance    & 1384      \\
4. Confuse      & 1224      \\
5. Resentment   & 1128      \\
6. Helplessness & 1089      \\
7. Anxiety      & 962       \\
8. Guilt        & 561       \\
9. Numbness     & 509       \\
10. Despair      & 461       \\
11. Fear         & 191       \\
\hline
\end{tabular}
\end{table}

\begin{table}
\centering
\caption{Distribution of speech segments by label count in a fine-grained emotion multi-label classification task.}
\label{tab:label_co_occurrence}
\begin{tabular}{ll} 
\hline
Number of labels & Segments  \\ 
\hline
1                & 5402      \\
2                & 2638      \\
3                & 1091      \\
4                & 289       \\
5                & 51        \\
6                & 2         \\
7                & 1         \\
\hline
\end{tabular}
\end{table}

\begin{figure}[!htbp]
\centering
\includegraphics[width=1\linewidth]{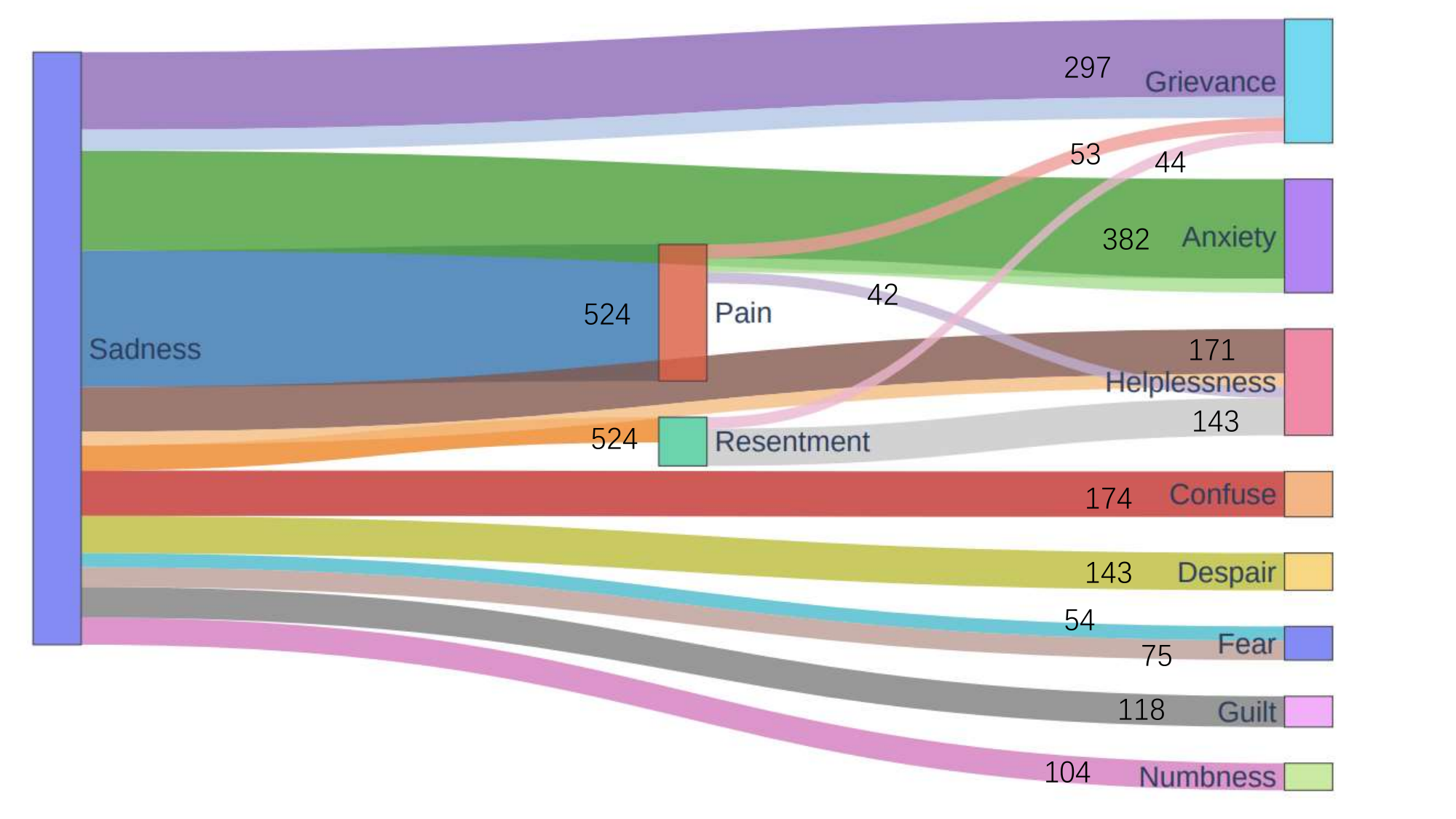}
\caption{Label co-occurrence relationships between categories in the fine-grained emotion multi-label classification task.}
\label{fig:label_co_occurrence}
\end{figure}

As shown in Table~\ref{tab:label_co_occurrence}, 57\% of the segments displayed a single emotion, though multiple emotions could co-occur, such as ``Sadness'' combined with ``Resentment'', or ``Sadness'' with ``Anxiety''. Some emotion labels, such as ``Sadness'' and ``Pain'' or ``Sadness'' and ``Grievance'', frequently co-occur, suggesting a possible intrinsic connection between these emotional states or a common likelihood of being triggered simultaneously in certain contexts. The relationships of label co-occurrence are further illustrated in Figure~\ref{fig:label_co_occurrence}.

\begin{figure}[!htbp]
\centering
\includegraphics[width=1\linewidth]{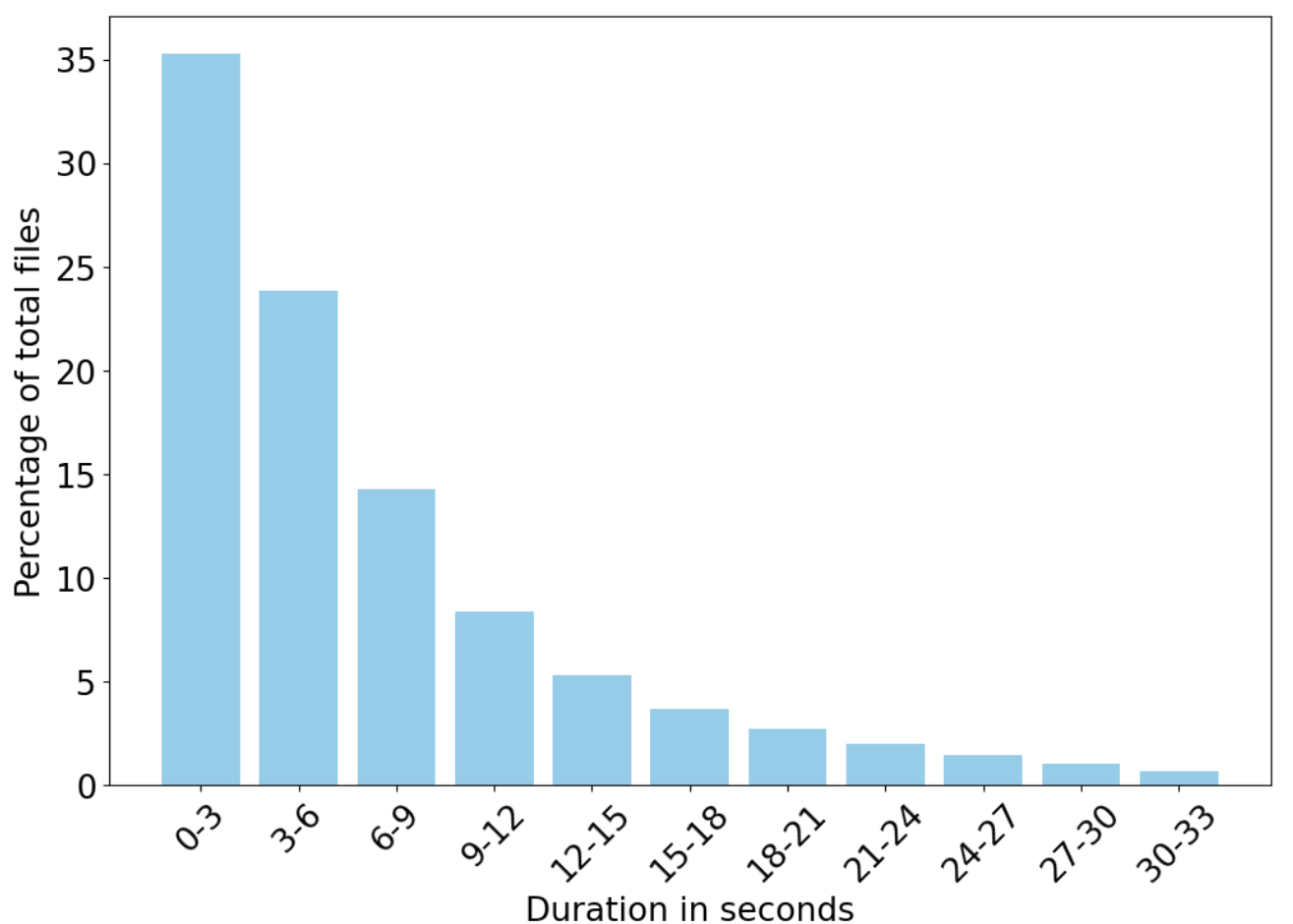}
\caption{Distribution of speech segment durations. Due to space constraints, we only show the top 11 categories with the highest percentages.}
\label{fig:segments_duration}
\end{figure}

The distribution of speech segments durations in our dataset is shown in Figure~\ref{fig:segments_duration}. The data show a prevalence of short speech segments, with 35.30\% of the samples ranging between 0 to 3 seconds and 23.82\% ranging from 3 to 6 seconds. Speech segments with durations between 6 to 15 seconds also represent a large proportion (27.96\%), which may indicate more complex expressions of emotion or more detailed descriptions of specific situations. Longer speech samples, exceeding 15 seconds, show a progressive decrease in frequency. In particular, there are only a few cases of samples that exceed 30 seconds. The average speech length in this dataset is 7.54 seconds, with the longest speech sample recorded at 63.61 seconds. The variability in speech duration, combined with the complexity of sentiment categories and the challenges posed by label attribution, complicates our analytical task.

\section{Methods} \label{sec:methods} 
We selected several large-scale pre-trained models: Wav2Vec 2.0~\cite{baevski2020wav2vec}, HuBERT~\cite{hsu2021hubert}, and Whisper~\cite{whisper_radford2023robust} which are based on the Transformer architecture~\cite{vaswani2017attention}. We conducted a comparative analysis using different versions of these models, varying in model size and training data. The details are as follows.
\subsection{Pre-trained feature extractor}
\paragraph{\textbf{Wav2Vec 2.0}} Wav2Vec 2.0~\cite{baevski2020wav2vec} is a framework designed for building automatic speech recognition (ASR) systems. This self-supervised learning framework learns feature representations directly from raw audio data. It encodes audio using a multi-layer convolutional neural network, then masks spans of the resulting latent speech representations, an approach similar to the training of language model like BERT~\cite{devlin2018bert}. These latent representations are subsequently fed into a Transformer~\cite{vaswani2017attention} model to build contextualized representations. The model is trained via a contrastive task, allowing it to learn representations solely from speech and then fine-tune on transcribed speech. This method can outperform the best semi-supervised methods while being conceptually simpler. Wav2vec 2.0 captures subtle changes in speech, such as pitch and speed, which are key factors in identifying emotions.

For our experiments, we used a version of Wav2Vec 2.0 that has been fine-tuned on multiple Chinese speech datasets, including Common Voice 6.1~\cite{commonvoice:2020}, CSS10~\cite{park2019css10}, and ST-CMDS~\cite{stcmds2017}. This fine-tuning enhances its relevance and effectiveness for the specific characteristics of Chinese speech.

\paragraph{\textbf{HuBERT}} 
HuBERT~\cite{hsu2021hubert} is a self-supervised learning model designed for speech processing tasks. It combines BERT's language model architecture and uses unique speech processing technology. The architecture of HuBERT is basically based on Transformers, which uses a self-attention mechanism to process sequence data. HuBERT is optimized for the characteristics of speech signals. Its training process consists of two stages. In the pre-training stage, the model is trained to predict hidden units in the speech signal. These hidden units are obtained by clustering analysis of pre-processed speech features, a process that mimics the masked language model task in traditional self-supervised learning. On the basis of pre-training, HuBERT uses more detailed labels or real downstream task labels for further training, such as the LibriSpeech~\cite{panayotov2015librispeech} data set, which can improve the model's adaptability and accuracy for specific speech tasks. HuBERT is particularly suitable for processing large-scale unlabeled speech data and can effectively extract useful information to support complex speech analysis tasks. HuBERT performs well when processing speech data with complex background noise, and can effectively identify the emotional content in speech.

In our experiments, we used the base version of HuBERT, which strikes a balance between model size and performance. Compared with the large version, it requires less computing resources for training and inference, which can help us quickly detect whether the HuBERT architecture is Suitable for our mission.

\paragraph{\textbf{Whisper}} Whisper is a pre-trained model designed for ASR and speech translation, trained on an extensive 680,000 hours of multilingual and multitask supervised data collected from the web~\cite{whisper_radford2023robust}. The architecture of Whisper is an end-to-end encoder-decoder Transformer~\cite{transformer_vaswani2017attention}. The process begins by splitting input audio into 30-second chunks and converted into a log-Mel spectrogram. This spectrogram is pass through an encoder, and a decoder is trained to predict the corresponding text caption, incorporating special tokens that enable the model to perform various tasks such as language identification, phrase-level timestamps, multilingual speech transcription, and speech-to-English translation.
Whisper demonstrates strong zero-shot performance, allowing it to generalize across many datasets and domains without the need of fine-tuning.

In our experiments, we utilized four versions of Whisper, including Whisper-small-Chinese-base, Whisper-small, Whisper-medium, and Whisper-large-v3. The Whisper-small-Chinese-base is a version of Whisper-small that has been fine-tuned on the Google Fleurs~\cite{conneau2023fleurs} ``cmn-hans-cn'' dataset, specifically optimized for Chinese language tasks. The Whisper-large-v3 model, compared to the standard large version, is trained for 2.5 times more epochs with added regularization to enhance performance.

\subsection{Fine-tuning on downstream tasks}
We used a pre-trained model to extract features from the speech segments, utilizing a hyperbolic tangent activation function and average pooling for the initial processing of the speech data. Following the feature extraction, we constructed a classifier designed to utilize the extracted speech representations as inputs for performing downstream tasks.

The classifier is composed of two linear layers interspersed with a dropout layer. In the forward propagation function of our model, the input features are first processed by the dropout layer, which randomly drops certain features. This is followed by a dense linear layer where the hyperbolic tangent activation function is applied to introduce non-linearity, enhancing the model’s capability to learn complex patterns in the data. Finally, the processed features pass through another linear layer, which outputs the classification scores for each category.

\section{Experiments and results} \label{sec:experiments} 

\subsection{Implementation details}

During model training, we divided the 105 speeches into training and test datasets at a ratio of 4:1. Specifically, 84 speeches were used for training and the remaining 21 for testing. 
We conducted 5-fold cross-validation within our training set to divide the 84 speeches into separate training and validation set. 
Note that, we split the data at the level of entire speeches rather than at the segment level to prevent data leakage.
In our experiments, we used accuracy, recall, and F1-score as evaluation metrics. Specifically, for the fine-grained emotion multi-label classification task, we report the performance as a weighted average to address the issue of data imbalance. 

The experiments were conducted using a NVIDIA 24GB RTX 4090 GPU. We set the batch size at 16 with a learning rate of 1e-4. Additionally, we implemented a gradient accumulation strategy, updating parameters every two accumulation steps. The model was trained for three epochs, utilizing half-precision training (fp16) to expedite the training process and decrease memory requirements. 

\subsection{Negative emotion recognition}
The experimental results for the negative emotion recognition task can be seen in Table~\ref{tab:exp_binary}.
\begin{table}[!h]
\centering
\caption{Performance of pre-trained models in a negative emotion recognition task.}
\label{tab:exp_binary}
\begin{tabular}{llll} 
\hline
Models                     & Accuracy & Recall  & F1       \\ 
\hline
Wav2Vec 2.0                & 76.97\%  & 76.96\% & 76.96\%  \\
HuBERT                     & 73.29\%  & 73.23\% & 73.23\%  \\
Whisper-small              & 74.94\%  & 74.92\% & 74.92\%  \\
Whisper-small-Chinese-base & 76.23\%  & 76.03\% & 76.03\%  \\
Whisper-medium             & 75.81\%  & 75.58\% & 75.58\%  \\
Whisper-large-v3           & 75.56\%  & 75.52\% & 75.52\%  \\
\hline
\end{tabular}
\end{table}

The evaluation of several pre-trained models for negative emotion recognition yielded close performance metrics, with Wav2Vec 2.0 slightly leading with an F1 score of 76.6\%. Wav2Vec 2.0's robust audio processing capabilities are particularly effective at interpreting emotional nuances in speech, which is critical for emotion recognition tasks. The Whisper Series models, which range from small to large, demonstrate that larger models do not necessarily yield better results in emotional recognition, as they all achieved approximately 75\% F1 scores. This suggests that improvements in model performance may be due to refined architectures or improved training approaches rather than simply scaling up the models.
Furthermore, the Chinese language fine-tuned version, Whisper-small-Chinese-base, showed only a marginal improvement, achieving a 1.11\% point increase in F1 score over the original Whisper-small version. In addition, HuBERT performed poorly in this task, suggesting that certain models may require specific adjustments to increase their effectiveness in complex emotional recognition scenarios.

\subsection{Fine-grained emotion multi-label classification}
The results of fine-grained emotion multi-label classification depicted in Table~\ref{tab:exp_fine} and Table~\ref{tab:exp_fine_details}.

\begin{table}
\centering
\caption{Performance of pre-trained models in the 11 classes fine-grained emotion multi-label classification task. The metrics of precision, recall, and F1 score are reported as weighted-averages.}
\label{tab:exp_fine}
\begin{tabular}{llll} 
\hline
Models                     & Precision & Recall  & F1       \\ 
\hline
Wav2Vec 2.0                & 56.50\%   & 23.98\% & 32.17\%  \\
HuBERT                     & 49.99\%   & 31.17\% & 38.03\%  \\
Whisper-small              & 50.06\%   & 31.45\% & 38.24\%  \\
Whisper-small-Chinese-base & 54.50\%   & 28.46\% & 36.30\%  \\
Whisper-medium             & 52.43\%   & 33.34\% & 40.34\%  \\
Whisper-large-v3           & 57.60\%   & 33.31\% & 41.74\%  \\
\hline
\end{tabular}
\end{table}

\begin{table*}
\centering
\caption{F1-score of pretrained models for each of the 11 categories of the fine-grained emotion multi-label classification task.}
\label{tab:exp_fine_details}
\resizebox{1\linewidth}{!}{
\begin{tabular}{llllllllllll} 
\hline
\diagbox{Models}{Categories} & Sadness & Pain    & Grievance & Confuse & Resentment & Helplessness & Anxiety & Guilt   & Numbness & Despair   & Fear     \\ 
\hline
Wav2Vec 2.0                  & 51.08\% & 31.49\% & 11.52\%   & 17.21\% & 11.40\%     & 2.14\%       & 2.56\%  & 0\%     & 6.56\%   & 7.69\%    & 0\%      \\
HuBERT                       & 53.41\% & 33.65\% & 22.71\%   & 24.66\% & 15.25\%    & 13.64\%      & 20.78\% & 18.46\% & 19.57\%  & 21.31\%   & 22.22\%  \\
Whisper-small                & 53.75\% & 34.06\% & 22.71\%   & 24.66\% & 15.25\%    & 13.59\%      & 20.78\% & 18.46\% & 19.57\%  & 21.31\%~~ & 22.22\%  \\
Whisper-small-Chinese-base   & 53.10\% & 34.91\% & 18.23\%   & 23.13\% & 12.56\%    & 2.07\%       & 29.41\% & 6.52\%  & 6.67\%   & 15.09\%   & 7.55\%   \\
Whisper-medium               & 53.53\% & 38.35\% & 27.31\%   & 27.15\% & 22.59\%    & 15.56\%      & 33.06\% & 18.6\%  & 11.63\%  & 24.00\%   & 27.78\%  \\
Whisper-large-v3             & 56.69\% & 38.78\% & 24.87\%   & 32.90\% & 15.14\%    & 4.74\%       & 35.98\% & 22.61\% & 30.14\%  & 25.81\%   & 35.44\%  \\
\hline
\end{tabular}
}
\end{table*}

Overall, the pretrained models showed limited performance in the fine-grained emotion recognition task, with a weighted F1 score of about 40\%. Notably, despite its underperformance in the negative emotion categorization task, HuBERT outperformed Wav2Vec 2.0 by 5.86\% points on the F1 score in this more detailed task. Similarly, the original version of Whisper-small outperformed its Chinese fine-tuned counterpart, Whisper-small-Chinese-base, by 1.94\% points on F1-score. These results underscore that different models may be better suited for specific types of task scenarios.
Despite the overall modest performance, the Whisper Series models still performed relatively well. However, the best performing model in the series, Whisper-large-v3, only achieved an F1 score of 41.74\%. This suggests that while large-scale training may offer some advantages, it does not automatically guarantee superior performance in complex tasks such as fine-grained emotion recognition.

Looking at the specifics of each category of classification, there is a direct correlation between model performance and sample size. For example, in the ``Sadness'' category, which contains the largest number of samples, all models achieve an F1 score above 50\%. The best performing model in this category, Whisper-large-v3, achieves an F1 score of 56.69\%. However, as the sample size decreases, there is a general trend of decreasing performance for all models.
It is particularly noteworthy that Whisper-large-v3 significantly outperforms other models in categories with fewer samples. In the ``Fear'' category, which has the smallest sample size of 191, Whisper-large-v3 outperforms its nearest competitor by 35.44\%. Similarly, in the ``Numbness'' category, which has 509 data points, it outperforms HuBERT's F1 score by 10.57\% points. These results highlight Whisper-large-v3's robust ability to perform well even in categories with smaller sample sizes, demonstrating that the model has a strong foundation and can achieve relatively good performance even with limited data.

\section{Discussion} \label{sec:discussion}
We selected prediction results from the Whisper-large-v3 model, the best performer in the fine-grained multi-label classification task, for error analysis, as detailed in Table ~\ref{tab:example_of_predictions}.

\begin{CJK*}{UTF8}{gbsn} 
\begin{table*}[!ht]
\centering
\caption{Examples of speech segment predictions by the best-performing Whisper-larve-v3 model in the fine-grained multi-label classification task. The table displays the original transcribed text alongside its English translation.}
\label{tab:example_of_predictions}
\resizebox{0.8\linewidth}{!}{
\begin{tabular}{llll} 
\toprule
  & Sentence                                                                                                                                                                  & Labels                                                                                    & Predictions                                                                   \\ 
\hline
A & \begin{tabular}[c]{@{}l@{}}[Chinese]~我和他们闹，一点用都没有，我甚至连哭都是自己一个。\\{[}Translation]~I argued with them, but it was of no use. I even cried alone.\end{tabular}                 & \begin{tabular}[c]{@{}l@{}}1. Sadness\\2. Pain\\6. Helplessness\\10. Despair\end{tabular} & \begin{tabular}[c]{@{}l@{}}1.~Sadness\\2. Pain\\6.~Helplessness\end{tabular}  \\ 
\hline
B & \begin{tabular}[c]{@{}l@{}}[Chinese]~很无奈，算了吧，不讲了吧，多费口舌。\\{[}Translation]~I'm very helpless, forget it, stop talking about it, it's a waste of words.\end{tabular}          & \begin{tabular}[c]{@{}l@{}}4. Confuse\\6. Helplessness\end{tabular}                       & None                                                                          \\ 
\hline
C & \begin{tabular}[c]{@{}l@{}}[Chinese]~我觉得是个坏事，让我害怕、担忧、坐立不安。\\{[}Translation]~I thought it was a bad thing and made me scared, worried and restless.\end{tabular}            & \begin{tabular}[c]{@{}l@{}}4.~Confuse\\6. Helplessness\\7. Anxiety\\11. Fear\end{tabular} & 7. Anxiety                                                                    \\ 
\hline
D & \begin{tabular}[c]{@{}l@{}}[Chinese]~我现在很怕碰我的小孩，我只要一碰他，他就跑掉。\\{[}Translation]~I am now afraid of touching my child. As soon as I touch him, he will run away.\end{tabular} & \begin{tabular}[c]{@{}l@{}}1.~Sadness\\2.~Pain\\7. Anxiety\\11. Fear\end{tabular}         & \begin{tabular}[c]{@{}l@{}}2.~Pain\\3. Grievance\\8. Guilt\end{tabular}       \\
\bottomrule
\end{tabular}
}
\end{table*}
\end{CJK*} 

Our analysis showed that the model often confuses closely related emotional states, such as ``Sadness'' and ``Despair'' or ``Pain'' and ``Grievance''. For example, in example (A), the category ``Despair'' is incorrectly omitted. The model also struggles with emotions that require a deeper understanding of context and nuance, as shown in example (B), where despite the caller's calm tone, emotions such as ``Confusion'' and ``Helplessness'' are present but not recognized by the model. The challenge extends to handling subtle emotions, as shown in example (C), where four complex emotions are present, but the model recognizes only one.

However, the model's performance is not uniformly poor. For example, in case (D), the described content suggests the emotion of ``Guilt'' which the model successfully predicts even though it was not annotated as such. This highlights the need to minimize annotator subjectivity and to better define and distinguish between emotions to improve data quality.

The dataset currently suffers from category imbalance, and future efforts should focus on expanding the training dataset with more diverse examples to improve categorization accuracy. In addition, the current model's approach, which is based on speech segments, lacks contextual links. Future development should aim to construct models that incorporate contextual data, allowing for nuanced recognition and linking of emotional states.

\section{Conclusion} \label{sec:conclusion} 

In this study, we explore the capability of large-scale pre-trained models for automatic emotion recognition by classifying emotions in speech data from the psychological support hotline of Beijing Huilongguan Hospital. We analyzed 20,630 segments from 105 callers to build negative emotion recognition models and identify 11 specific negative emotions. While these models achieved a commendable F1 score of 76.96\% in binary classification tasks, their performance in multi-label emotion recognition tasks was less impressive, with the best model achieving a weighted F1 score of only 41.74\%.
Our findings highlight the potential of using deep learning techniques and large datasets to improve the effectiveness of mental health hotline services. Future efforts will aim to improve the model's ability to detect fine-grained emotions. This will involve increasing the number of training samples and refining the model structure to increase accuracy across different emotion categories. Since the model's performance on specific emotions is strongly influenced by the amount of training data, increasing the variety of data samples and refining the model structure will be crucial for improving overall performance.
Overall, our research contributes to the growing understanding of emotion recognition in the context of psychological crisis, highlighting both achievements and areas for further development.

\bibliographystyle{IEEEtran}
\bibliography{ref}
\end{document}